# Web-based Semantic Similarity for Emotion Recognition in Web Objects

Valentina Franzoni†*, Giulio Biondi*, Alfredo Milani*°, Yuanxi Li°

†Department of Computer, Control, and Management Engineering, Sapienza University of Rome, Italy
franzoni@dis.uniroma1.it
*Department of Mathematics and Computer Science, University of Perugia, Italy
°Department of Computer Science, Hong Kong Baptist University, Hong Kong, China

**ABSTRACT**

In this project we propose a new approach for emotion recognition using web-based similarity (e.g. confidence, PMI and PMING). We aim to extract basic emotions from short sentences with emotional content (e.g. news titles, tweets, captions), performing a web-based quantitative evaluation of semantic proximity between each word of the analyzed sentence and each emotion of a psychological model (e.g. Plutchik, Ekman, Lovheim). The phases of the extraction include: text preprocessing (tokenization, stop words, filtering), search engine automated query, HTML parsing of results (i.e. scraping), estimation of semantic proximity, ranking of emotions according to proximity measures. The main idea is that, since it is possible to generalize semantic similarity under the assumption that similar concepts co-occur in documents indexed in search engines, therefore also emotions can be generalized in the same way, through tags or terms that express them in a particular language, ranking emotions. Training results are compared to human evaluation, then additional comparative tests on results are performed, both for the global ranking correlation (e.g. Kendall, Spearman, Pearson) both for the evaluation of the emotion linked to each single word. Different from sentiment analysis, our approach works at a deeper level of abstraction, aiming at recognizing specific emotions and not only the positive/negative sentiment, in order to *predict emotions* as semantic data.

## 1. INTRODUCTION

The interest on *emotion tagging and recognition* grew in last decade in several areas of research, from social network analysis [11][12][34] to recommender systems [32], from image [6][10][33] to music [36] recognition, to semantic context generation [9][35]. With the pervasive diffusion of social digital interactions [21], *big data analysis* has become a critical issue, as an increasing amount of economical, personal and physical transactions is digitally mediated, recorded and monitored, and can be used to predict and influence the way people or entities interact, work and, in general, behave. One of the mainstream approaches to semantic annotation and retrieval is *sentiment analysis*, which goal is to classify sentiments about objects as positive/negative/neutral. Emotions contained in *multimedia* data, i.e. emotions that are expected to be elicited by the object, although, have a greater emotional value than the one that can be expressed by sentiments. Since the middle of the Twentieth century, psychologists and sociologists put a great effort to set some general models of emotions and to make *quantitative tests* on humans, in order to decide which emotions can be included in general models and which of them are measurable. While sentiments include information about positive and negative states, emotions instead are different concepts. Sentiments can be associated with events or objects, relying on previous memories about other events related to the object. Basic emotions, or affective responses, include impulses that will be the same for every human being, and can thus be generalized.

In order to express emotional categories, some main emotions can be included in a model (e.g. in the *Ekman*, *Plutchik*, *Lovheim* models) [32]. Each model can show emotions in different ways and with different levels, but the basic emotions will not themselves differ. Besides the concepts of these basic emotions, words in a language can be used to express them, using *nouns* or *adjectives*, or even *adjective pairs* or *expressions* (e.g. groups of terms). We represent each *object as a vector of emotions*, where each position represents a *weighted value* [18] for the emotions included in the model under representation. In this work, three models are considered (Plutchik, Ekman and Lovheim), and in general the code allows to add other models as vector spaces: objects, emotions and models of emotions are, in fact, parametrized in the implementation code.

Aiming at defining a framework for semantic emotional context generation, a suitable knowledge source is needed, to provide semantic relationships among concepts, together with a suitable technique for extracting new concepts, relevant and consistent with the seed context. If, on one hand, existing semantic models using ontologies [1] are expressive enough, on the other hand they find their basic limitation on the evolution and management of ontological models and annotated content [8], which are not taken into account in the model itself [2]. It results in a lack of automation capabilities and evolutionary maintenance that is highly relevant, especially for real-time applications. Using web-based proximity measures, the limitation of not having evolutionary results, updated in real time, can be overcome [4][30]. Several subroutines are used in each step, in order to accomplish the tasks of the related phase. In the following paragraphs, the main processing phases of a web object are described in detail. The process can be repeated for every object/sentence in the data set, or data set subset. Being able to support and adapt to different combinations of *model assets* (i.e. basic emotion models, proximity measures, search engines, and data sets), our implementation structure is flexible, and new assets can be added without making any change to the code, by adding JSON files only, containing such files in the appropriate folders. This makes the code *reusable* and *robust* to changes.

## 2. WEB-BASED PROXIMITY

Search Engine (SE)-based measures [6][22][37] evaluate the proximity of two terms using statistics extracted from search engine queries. Several advantages of using SE-based measures can be pointed out: the statistical data can be *easily extracted* by querying the SE, the minimum requirement is that SEs provide the number of occurrences and co-occurrences of terms, the results returned by a SE *change over the time* and are *continuously updated*. The proximity measure will then reflect the *collective knowledge*, at the time of the query, of the community producing the documents/multimedia indexed by the engine. Any search engine (e.g. Bing, Google, Yahoo) [13-16], and any web-based semantic proximity measure [3][5] (e.g. Confidence, PMI, NGD, PMING Distance), can be used in this approach, where normalized distances are *1-proximity*:

***Confidence (CF)*** [5] is an asymmetric statistical measure, used in rule mining for measuring trust in a rule $X \rightarrow Y$ i.e., given the number of transactions that contain $X$, then the *Confidence* in the rule indicates the percentage of transactions that contain also Y.

$$confidence(x \rightarrow y) = \frac{P(x,y)}{P(x)} = \frac{f(x,y)}{f(x)}$$

From a probabilistic point of view, confidence approximates the conditional probability, such that

$$confidence(x \rightarrow y) = P(y|x) = \frac{P(x|y)P(y)}{P(x)}$$

where $\frac{P(x|y)}{P(x)} = \frac{P(y|x)P(x)}{P(y)} = \frac{P(y|x)}{P(0y)}$ is the *a priori* likelihood function for statistical inference.

***Pointwise Mutual Information (PMI)*** [19][20][24] is a point-to-point measure of association used in statistics and information theory. Mutual information between two particular events $w_1$ and $w_2$, in this case two words $w_1$ and $w_2$ in webpages indexed by a search engine, is defined as:

$$PMI(w_1, w_2) = log_2 \frac{P(w_1,w_2)}{P(w_1)P(w_2)} \quad (4)$$

$PMI(w_1, w_2)$ is an approximate measure of the quantity of information provided by the occurrence of the event/term $w_2$ about the occurrence of the event/work $w_1$. PMI is a good measure of independence while represent a bad measure of dependence, since the dependency score is related to the frequency of individual words.

***Normalized Google Distance (NGD)*** [7][26] has been introduced as a measure of semantic relation, based on the assumption that similar concepts $x$ and $y$ occur together in a large number of documents in the Web i.e., the frequency of documents returned by a query on a search engine $S$ approximates the distance between related semantic concepts:

$$NGD(x,y) = \frac{max\{\log f(x), \log f(y)\} - \log f(x,y)}{\log M - min\{\log f(x), \log f(y)\}}$$

$f(x)$, $f(y)$ and $f(x,y)$ are the cardinalities of results returned by $S$ for the query on $x$, $y$, $x \wedge y$ respectively, and $M$ is the number of pages indexed by $S$, or a value reasonably greater than $f(x)$ for each possible $x$.

***PMING Distance (PMING)*** [3][37] consists of NGD and PMI locally normalized, with a correction factor of weight $\rho$, which depends on the differential of NGD and PMI.

More formally, the PMING distance of two terms $x$ e $y$ in a context $W$ is defined, for $f(x) \geq f(y)$, as a function $PMING: W \times W \rightarrow [0,1]$

$$PMING(x,y) = \rho \left(1 - \log \frac{f(x,y)M}{f(x)f(y)\mu_1}\right) + (1 - \rho) \left(\frac{\log f(x) - \log f(x,y)}{(\log M - \log f(y))\mu_2}\right)$$

where $\mu_1$ and $\mu_2$ are constant values which depend on the context of evaluation, defined as $\mu_1 = \max PMI(x,y)$ and $\mu_2 = \max NGD(x,y)$ with x, y $\in$ W.

Previous experiments [3][37] show that PMING is one of the measures with the best overall performance for clustering and classification of concepts. *Conf()* (i.e. *Confidence*) provides the best results for shortest semantic path finding in Wikipedia [4], because *confidence* better reflects the *explanatory relationships*, with respect to other web-based similarity measures.

## 3. MODELS OF EMOTIONS

In this work three models of emotions [32] are considered:

$E_{Ekman.}$=[*anger, disgust, fear, joy, sadness, surprise*]

$E_{Plutchik}$=[*anger, anticipation, disgust, fear, joy, sadness, surprise, trust*]

$E_{Lovheim}$=[*anger, disgust, distress, fear, interest, joy, shame, surprise*]

These models include the basic emotions for human behavior, but also other models can be easily included as parameters in the code. Some preliminary tests have been made to check if adjectives have a better performance than nouns. Being terms queried directly on a web-based search engine, stemming is not needed in the preprocessing phase.

## 4. STATISTIC SCALES FOR BEHAVIOR

Many statistic measurement scales [31] exist for studying human behaviors. The most common ones are the *Thurstone* [28] and *Guttman* [27] scales, that have main disadvantages for human behavioral evaluation, and *Likert* [25] and *semantic differential* [23] scales, which instead are among the most used for behavioral measurement and for diagnostic tests by human evaluation. In the following paragraph, a brief description of these scales will be given. In general, scales for measuring the behaviors are by rule built of sets of items to which a human is asked to give a feedback, which can be *positive* or *negative*. The underlying hypothesis of this approach is that through this feedback it should be possible to measure human behavior with respect to an event or topic along a *continuum* of goodness/badness of each evaluation point of the scale. A deeper discussion of what is a behavior, and if measuring it makes sense, are not within the scopes of this work, and can be studied in specialized books in the psychology area.

If in some scales of intervals, e.g. temperature scales, is easy to say that a constant unit exists, and that a "zero" arbitrarily chosen can be established, in the empirical system of behavioral models this is not trivial. Also, it is not trivial to proof that the unit does not change during the *continuum*. It is although possible to perform tests on the distributions of the features, with the aim of elaborating and interpreting results. If it is possible to identify an element of null intensity (i.e. a "zero") in the system, then *the system will have all the properties of a numeric system*, and transformation rules can be used, including proximity/distance among items.

Hybrid methods can use a combination of different scales, in order to evaluate a set of items/features.

## 5. THE PROPOSED FRAMEWORK

In this project, we aim to extract basic emotions from short emotionally reach sentences (e.g. news titles, tweets, captions) performing a web-based quantitative evaluation of semantic proximity between each word of the analyzed sentence and each emotion of a chosen psychological model.

The phases of the extraction include:

1. *text preprocessing* (tokenization, stop words, filtering)
2. *search engine automated query*
3. *scraping of results* (i.e. parsing of the frequency of documents returned by the search engine)
4. estimation of the web-based *semantic proximity*
5. *ranking of emotions*.

The main idea is that *it is possible to generalize semantic similarity under the assumption that similar concepts co-occur in documents indexed in search engines* [37], and therefore also emotions can be generalized in the same way, through those tags or terms that express them in a language. We use web-based semantic proximity measures (e.g. confidence, PMI and PMING) to evaluate the similarity of each term of a set related to a Web object (e.g. image, description, comment), analyzing the set with respect to each emotional word of a psychology model of emotions. A ranking of emotions is then created, based on semantic proximity, and an aggregation of results is performed, so that every sentence can be represented as a vector of weighted emotions with the *Vector Space Model (VSM)*.

More formally, our is a general model for emotion ranking based on semantic proximity measures proposed in this work. The proposed semantic model is characterized by a proximity measure $\eta$ and a basic emotions model $E=\{e_1,…,e_n\}$. Each term $t$ is associated to an emotion vector $v_t=[v_{t,1},…,v_{t,n}]$ where each dimension $I$ correspond to emotion $e_i$ in $E$, and its value $v_{t,i}$ is given by the normalized proximity measure $\eta(t,e_i)$ between the term $t$ and the emotion $e_i$. In other words, each term corresponds to a point in the VSM of the emotions in $E$. Emotion rankings of web objects are then obtained by aggregating the values of the emotion vectors associated to the constituent terms of its textual description.

In our framework, various basic emotions model and proximity measures have been experimented, $E \in \{E_{Plutchik}, E_{Ekman}, E_{Lovheim}\}$ and $\eta \in \{PMI, NGD, Conf, PMING\}$.

## 5.1 Data Collection

In the first phase, a script will gather the data from a search engine. The search engine is a *parametrized asset*: three search engines (Bing, Google, Yahoo) [13-15] are available and can be chosen by a dropdown menu in our graphic user interface, but the code is suitable to easily enlarge the set of compatible search engines. A specific search engine can thus be called as a parameter of the main search method.

*5.1.a Preprocessing.* After loading the other assets, i.e. the *sentence* and the *emotion model*, a preprocessing of the sentence is performed, using parsing functions and NLP methods. A subroutine takes care of *tokenizing* and deleting all the tokens that are not needed (i.e. *stop words*), thus returning an array containing only the relevant words from the point of view of the emotional semantic content. In this work, in fact, words are taken as tokens, while some syntactical elements, such as pronouns and articles, are discarded, because they do not have by definition a relevant emotional meaning. This step has been decided after preliminary experiments, with the aim of *filtering* the information set to search, thus *reducing the time-based complexity* of the search phase. During the filtering step, all the items belonging to at least one of the following categories are removed:

- *Stop-words* contained in the English dictionary, part of the *all-corpora* package of the *Python NLTK Toolkit* library [29]. The whole "stopwords" dictionary contains common stop-words from 11 languages. NLTK is a leading open source platform for building Python programs to work with human language data. It provides easy-to-use interfaces to over 50 corpora and lexical resources such as WordNet [1].

- *Ordinal numbers*, detected using a regular expression in the language of evaluation (e.g. English).

- *Cardinal numbers*, including any sequence of numeral characters, each one in the *0-9* range.

- For the English and Italian language, *words with length≤3*, which include all articles and prepositions.

These steps are necessary both to avoid performing unnecessary queries and to keep only relevant data, from the point of view of emotional content, to be used in the analysis phase.

*5.1.b Automated Search.* When the list of words to search has been created, the actual *automated query* starts. For each emotion in the chosen model, a web search is performed on the chosen search engine. For automatic web searches, a web driver has been used, called *Selenium* [17], that allows automation of browser control, thus exploiting - step by step - the same actions that a human user will perform. In particular, the script will open a browser window, load the search engine page, type - letter by letter – and submit the query, simulating as precisely as possible a human search.

*5.1.c Scraping.* At the end of the page loading, the source code of the page is accessible through Selenium. A dedicated subroutine will then *scrape the webpage* to extract the number of results returned, using appropriate methods for each search engine. The web browser window will then be closed. It is important to note that each time Selenium opens a new window, it will use a totally new *blank profile*. This ensures that search results are not affected by the user's personal or history data, much like using the *Incognito* mode available on most of modern browsers.

Our script can identify bans and manage results, discarding these filtered ones: in case of numbers below the threshold, or no results at all, a *ban warning* will be displayed and the script will wait some *random delay* before performing again the query, until a correct result is obtained. Another random delay is introduced between every successful search, to *mimic a real user's behavior*. The same process will be repeated for each single word in the sentence and for each word-emotion and emotion-word pair. At the end the phase, three JSON files will contain emotion occurrences, words occurrences, and occurrences of pairs word-emotion: let $K$ be the number of emotions in the model and W the number of words after preprocessing; then, the total number of queries will be $T=K+W+K*W*2$.

## 5.2 EMOTION RANKING

Since PMING, in our implementation, needs the maximum value of PMI and NGD among all the pairs, its values are firstly initialized to *zero* (0) and then a second loop goes again through all the pairs, calculating PMING values. In the implementation hereby proposed, therefore, PMING calculation will be locked to the calculation of PMI and NGD, which is the less complex way to calculate PMING if PMI and NGD values are available. At the end of the calculation, a dictionary is dumped in a JSON file, which contains the values of all the measures calculated for each word and each emotion, i.e. for each sentence/tag set. The script produces some CSV files used to present data in an aggregate, easily readable (by humans) and re-usable (from computers) format. The files will contain a ranking of the emotions (i.e. the vector space) locally for each word, and globally for each sentence. Furthermore, a further result file containing all the sentence-level measures is provided as a data set.

The *Average* and *Max* functions are then calculated per column, as well as the rankings induced by the aggregated measure among emotions.

## 6. EXPERIMENTS AND RESULTS

The aim of preliminary experiments is at evaluating the contribution of our model, in different scenarios. Choosing the data set of news title is a good way to face with several different *contexts* and *topics*. Any data set of web objects, which can be represented as a *vector space of emotions*, is suitable: e.g., a good data set could be collected by social networks, such as Twitter. We choose the *SemEval-2007* data set [38], which gives knowledge of *human evaluations* on a Linkert scale, which can be used as a ground truth. Since the evaluation is carried out on each title, and on each relevant word is extracted after the preprocessing phase, results can be easily comparable with other approaches, which can be a future test. SemEval-2007 includes *250 news titles and documents from the Web*, and their ground truth with *human evaluations of emotions* of the *Ekman model*.

In Table 1, a sample of results is shown: in each row, the vector space for the sentence is shown. Only few samples are given, because of space limit: the sample is representative of results.

**Table 1: Example of the Emotion Vector Spaces for PMI**

| ID | Anger | Disgust | Fear | Joy | Sadness | Surprise |
|---|---|---|---|---|---|---|
| 12 | 0,173 | 0,134 | 0,171 | 0,153 | 0,192 | 0,175 |
| 56 | 0,187 | 0,164 | 0,168 | 0,126 | 0,191 | 0,161 |
| 127 | 0,211 | 0,195 | 0,178 | 0,114 | 0,151 | 0,150 |
| 168 | 0,197 | 0,195 | 0,198 | 0,115 | 0,165 | 0,128 |
| 193 | 0,173 | 0,160 | 0,165 | 0,135 | 0,196 | 0,169 |
| 247 | 0,179 | 0,184 | 0,189 | 0,081 | 0,207 | 0,156 |

## 6.1 Evaluation Criteria

In order to evaluate the quality of results, the following correlation coefficients, recalled directly by a Python library, have been computed between the emotional rankings induced by our experiments and the SemEval-2007 ground truth: *Spearman ρ*, *Kendall τ*, *Pearson r*, plus a sentiment evaluation.

**Table 2: correlation values for the emotion vector space**

| ID | Sentence Text | Pearson | Spearman | Kendall |
|---|---|---|---|---|
| 12 | Nicole Kidman asks dad to help stop husband's drinking | 0,931 | 0,811 | 0,690 |
| 56 | Marine killed in fighting west of Baghdad | 0,678 | 0,880 | 0,745 |
| 127 | German paper shows soldiers desecrating skull | 0,740 | 0,869 | 0,690 |
| 168 | Growing Unarmed Battalion in Qaeda Army Is Using Internet to Get the Message Out | 0,942 | 1 | 1 |
| 193 | Gunman fine before shooting | 0,922 | 0,819 | 0,745 |
| 247 | Gunmen kill 11 in Iraq TV raid | 0,805 | 0,811 | 0,690 |

At sentence level, we can provide a comparison with respect to the data set ground truth. This comparison shows that similarity indexes are in the worst case around 67%, with a remarkable 82% on average, which is a promising result for a web-based proximity approach. In Table 2, the correlation between our results and the ground truth is shown for the same sample of Table 1 from the data set, where for each sentence the ranking correlation values are shown using Kendall τ, Spearman ρ, and Pearson *r*. Our model is also useful as a basis for clearly readable *visualizations of the emotional content* of the semantic object. In Fig. 1, an example of visualization for E=E$_{Ekman}$ and η=PMI is shown, using a radar graph on the sentence n.247 of SemEval-2007 (i.e. the news title: "*Gunmen kill 11 in Iraq TV raid*"). On the top legend, terms obtained after preprocessing the title are visible; on the radar, the emotional load is shown.

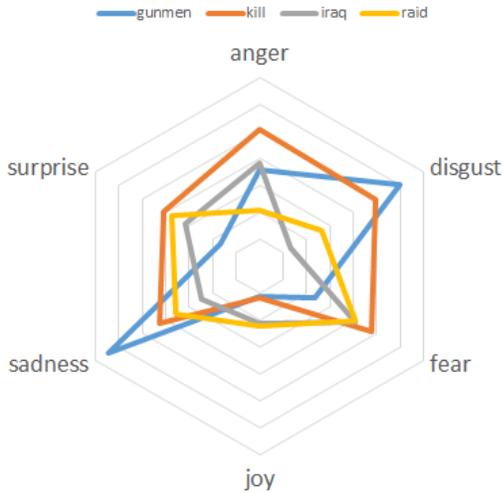

**Fig. 1: Visualization of title n.247 using a Radar Graph for E=E$_{Ekman}$.and η=PMI**

## 7. CONCLUSIONS

In this work, the role of web-based semantic similarity is modeled and experimented for emotion recognition in web objects. Besides the prominently linguistic tasks, other semantic dimensions are worth to be investigated.

We proposed and experimented a model for emotion ranking, based on semantic proximity measures, e.g. Confidence, PMI, PMING. The emotions of textual descriptions/tags of a web object are obtained by composing the emotion vectors associated to the constituent terms. Each dimension of the vector of emotions corresponds to an emotion, and the component value is given by the normalized proximity measure between the term and the emotion. More formally, each term corresponds to a point in the vector space model of the emotion.

In the proposed general framework, various basic emotions model and proximity measures have been experimented: $E \in \{E_{Plutchik}, E_{Ekman}, E_{Lovheim}\}$ as models of emotions, and $\eta \in \{NGD, Confidence, PMI, PMING\}$ as proximity measures.

Results show that both on a single-word level (accuracy, F1) and on a sentence-based level (Kendall τ, Spearman ρ, and Pearson *r*) our model provides interesting results. Future experiments are worth, to show that our web-based approach can give promising results on general topics.

Different from sentiment analysis, our approach works at a deeper level of analysis, aiming to recognize specific emotions and not only the positive/negative sentiment, to predict emotions as semantic data. Such affective information can be used in various personalized systems like recommender systems, human-machine interfaces, social robots, to use or create behavior models.

## 8. Aknowledgements

The authors thank Mr. Ka Ho TAM, graduate student of Hong Kong Baptist University, for the useful discussion and support.